\documentclass[letterpaper]{article} % DO NOT CHANGE THIS
\usepackage{aaai2026}  % DO NOT CHANGE THIS
\usepackage{times}  % DO NOT CHANGE THIS
\usepackage{helvet}  % DO NOT CHANGE THIS
\usepackage{courier}  % DO NOT CHANGE THIS
\usepackage[hyphens]{url}  % DO NOT CHANGE THIS
\usepackage{graphicx} % DO NOT CHANGE THIS
\usepackage{natbib}  % DO NOT CHANGE THIS AND DO NOT ADD ANY OPTIONS TO IT
\usepackage{caption} % DO NOT CHANGE THIS AND DO NOT ADD ANY OPTIONS TO IT
\usepackage{amsfonts}
\usepackage{amsmath}
\usepackage{multirow} 
\usepackage{booktabs}
\usepackage{algorithm}
\usepackage{algorithmic}

\usepackage{newfloat}
\usepackage{listings}
\DeclareCaptionStyle{ruled}{labelfont=normalfont,labelsep=colon,strut=off} % DO NOT CHANGE THIS
\floatstyle{ruled}
\newfloat{listing}{tb}{lst}{}
\floatname{listing}{Listing}
\title{Hypergraph Mamba for Efficient Whole Slide Image Understanding}
\author{
    Jiaxuan Lu\textsuperscript{\rm 1}\equalcontrib,
    Yuhui Lin\textsuperscript{\rm 2}\equalcontrib,
    Junyan Shi\textsuperscript{\rm 2}\equalcontrib,
    Fang Yan\textsuperscript{\rm 1},
    Dongzhan Zhou\textsuperscript{\rm 1},\\
    Yue Gao\textsuperscript{\rm 3},
    Xiaosong Wang\textsuperscript{\rm 1}\thanks{Corresponding author}
}
\affiliations{
    \textsuperscript{\rm 1}Shanghai Artificial Intelligence Laboratory\\
    \textsuperscript{\rm 2}Xi’an Jiaotong–Liverpool University\\
    \textsuperscript{\rm 3}Tsinghua University\\
    lujiaxuan@pjlab.org.cn, wangxiaosong@pjlab.org.cn
}

\usepackage{bibentry}
\begin{document}

\maketitle

\begin{abstract}
% Whole Slide Images (WSIs) in histopathology present a significant challenge for large-scale medical image analysis due to their high resolution, large size, and complex tile relationships. Existing Multiple Instance Learning (MIL) methods, such as Graph Neural Networks (GNNs) and Transformer-based models, face limitations in scalability and computational cost. To bridge this gap, we propose the WSI-HGMamba framework, which synergistically combines the relational modeling strengths of Hypergraph Neural Network (HGNN) with the efficiency of Mamba, the State Space Model designed for sequence learning. The proposed HGMamba block integrates Message Passing, Hypergraph Scanning \& Flattening, and feature aggregation via a Bidirectional State Space Model (Bi-SSM), achieving Transformer-level performance with $7\times$ fewer FLOPs. By leveraging the complementary strengths of lightweight HGNN and Mamba, the WSI-HGMamba framework delivers a scalable solution for large-scale WSI analysis, offering both high accuracy and computational efficiency for slide-level classification.
Whole Slide Images (WSIs) in histopathology pose a significant challenge for extensive medical image analysis due to their ultra-high resolution, massive scale, and intricate spatial relationships. Although existing Multiple Instance Learning (MIL) approaches like Graph Neural Networks (GNNs) and Transformers demonstrate strong instance-level modeling capabilities, they encounter constraints regarding scalability and computational expenses. To overcome these limitations, we introduce the WSI-HGMamba, a novel framework that unifies the high-order relational modeling capabilities of the Hypergraph Neural Networks (HGNNs) with the linear-time sequential modeling efficiency of the State Space Models. 
At the core of our design is the HGMamba block, which integrates message passing, hypergraph scanning \& flattening, and bidirectional state space modeling (Bi-SSM), enabling the model to retain both relational and contextual cues while remaining computationally efficient. Compared to Transformer and Graph Transformer counterparts, WSI-HGMamba achieves superior performance with up to $7\times$ reduction in FLOPs. Extensive experiments on multiple public and private WSI benchmarks demonstrate that our method provides a scalable, accurate, and efficient solution for slide-level understanding, making it a promising backbone for next-generation pathology AI systems.
% The proposed HGMamba block combines Message Passing, Hypergraph Scanning \& Flattening, and feature aggregation through a Bidirectional State Space Model (Bi-SSM), achieving Transformer-level performance while reducing FLOPs by a factor of 7. 
% Taking advantage of the complementary attributes of the HGNN and Mamba, the WSI-HGMamba framework provides a scalable solution for comprehensive WSI analysis, delivering superior accuracy and computational efficiency for slide-level understanding.
\end{abstract}


\section{Introduction}
Pathology AI diagnosis plays a pivotal role in advancing medical image analysis, particularly within the domain of histopathology. At the center of this progress are Whole Slide Images (WSIs), the ultra-high-resolution scans of entire tissue slides which encode an enormous amount of morphological and pathological information. Each WSI typically contains tens of thousands of image tiles, capturing subtle yet critical spatial cues such as cell distribution, tissue architecture, and microenvironmental interactions. Accurately interpreting these slides is essential for tasks such as cancer subtyping, prognosis estimation, and biomarker prediction. However, the sheer scale and complexity of WSIs pose significant computational and modeling challenges for conventional deep learning approaches.
% Pathology AI diagnosis plays a critical role in medical image analysis, particularly with Whole Slide Images (WSIs) in histopathology. WSIs are high-resolution images generated by scanning tissue samples at a microscopic level, capturing a vast amount of information about the tissue structure, morphology, and pathology. These images contain large numbers of tiles ranging from thousands to tens of thousands, which require complex modeling to capture the intricate relationships between them for accurate slide-level classification.

Multiple Instance Learning (MIL)~\cite{li2021dual,tang2023multiple,yao2020whole,lu2024pathotune} has emerged as a standard paradigm for understanding WSI. 
In MIL, the WSI is represented as a bag of instances (\textit{i.e.}, tiles), and labels are only available at the slide level as weak supervision. This formulation has catalyzed the development of various instance-level modeling strategies, including attention-based mechanisms~\cite{zheng2020diagnostic,weitz2021investigation,zhang2024attention}, Graph Neural Networks (GNNs)~\cite{zhao2020predicting,chen2021whole,chan2023histopathology,yu2024dualgcn}, Transformers~\cite{chen2021multimodal,li2024rethinking}, and Graph Transformers~\cite{zheng2022graph,shirzad2023exphormer,shi2024integrative}. Additionally, Hypergraph Neural Networks (HGNNs) have gained increasing traction due to their robust ability to model high-order relationships that go beyond simple pairwise connections of GNNs. By connecting multiple nodes through hyperedges, HGNNs can naturally model complex structure correlations such as glandular patterns, tumor-infiltrating lymphocytes, or vascular networks within a single computational unit~\cite{di2022generating,gao2024hypergraph,han2025inter}. 
% have shown superior capabilities in modeling complex and high-order relationships among tiles in WSIs, which connect multiple nodes simultaneously through hyperedges, capturing richer contextual dependencies across tissue structures with relatively low computational cost~\cite{di2022generating,gao2024hypergraph,han2025inter}. 

\begin{figure}[t]
\includegraphics[width=\linewidth]{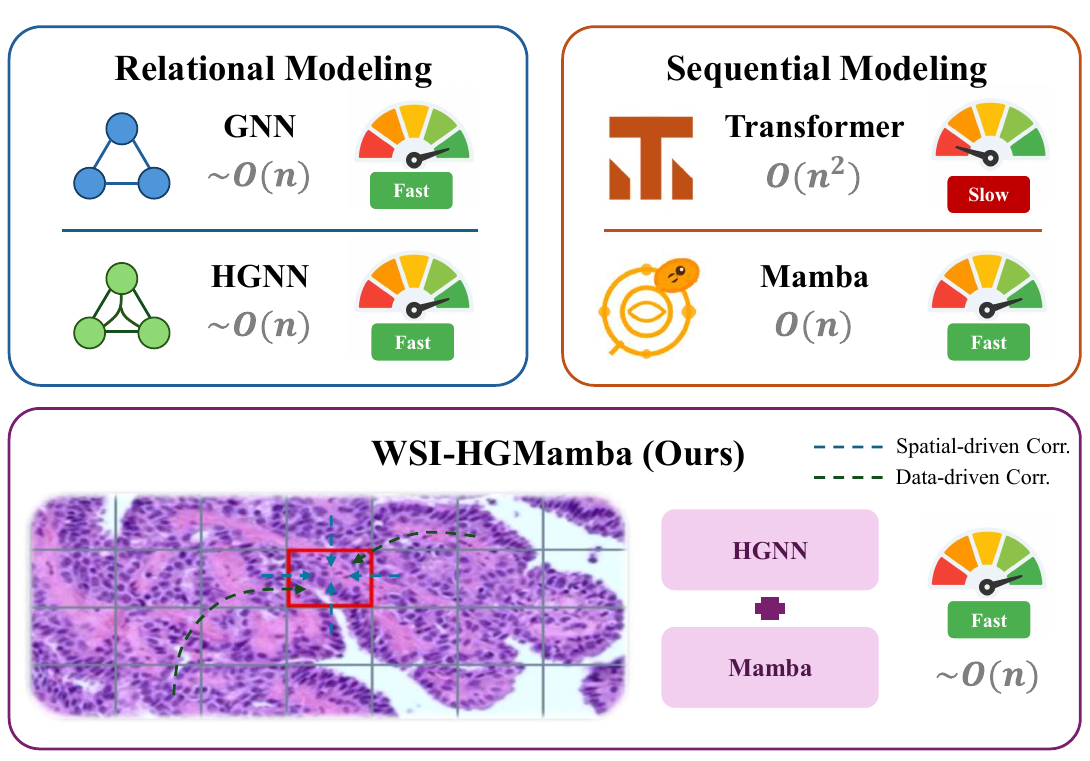}
\caption{Compared to existing relational and sequential modeling methods, WSI-HGMamba jointly captures high-order spatial correlations and sequential contextual information in ultra-high-resolution WSIs, while maintaining low computational complexity.} \label{first_page}
\end{figure}

Parallelly, sequential modeling methods have also proven highly effective by capturing long-range dependencies through global attention mechanisms. In histopathology, Transformers have enabled breakthroughs in fine-grained tile-to-tile sequential modeling~\cite{chen2021multimodal,shao2021transmil,li2024rethinking,wang2022transformer,gao2024transformer}. Graph Transformer methods further combine the high-order relational modeling of GNNs with the expressive power of Transformers~\cite{zheng2022graph,shirzad2023exphormer,shi2024integrative}.
However, their quadratic time and memory complexity $O(n^2)$ limits their scalability, especially in large WSIs, which can contain upwards of 50,000 tiles~\cite{yang2024mambamil,xu2024whole}.
% However, their quadratic complexity $O(n^2)$ of self-attention limits their scalability, especially in large WSIs~\cite{yang2024mambamil,xu2024whole}. 

% GNN-based models excel in modeling the relationships within WSIs, supporting high node counts with relatively low complexity. On the other hand, Transformer-based methods have demonstrated strong performance by capturing long-range dependencies through self-attention mechanisms. However, the quadratic complexity $O(n^2)$ of self-attention limits their scalability, especially in large WSIs~\cite{yang2024mambamil,xu2024whole}. Graph Transformer models aim to bridge the gap by combining the strengths of GNNs and Transformers, offering both relational modeling and powerful learning capacity. Despite their potential, their high computational cost remains a significant hurdle for large-scale WSI analysis.

The State Space Model, \textit{e.g.}, Mamba~\cite{mamba,mamba2,xing2024segmamba,lin2024event,behrouz2024graph,lin2025mv,liu2025msgm}, designed for efficient sequential learning, offers a compelling alternative. Designed to capture long-range dependencies with linear complexity $O(n)$, SSMs shift away from attention-based architectures and instead use dynamic state propagation mechanisms. Recent works~\cite{xing2024segmamba,lin2024event,liu2025msgm} have demonstrated that Mamba can outperform Transformers in efficiency while retaining competitive accuracy in sequential modeling. However, Mamba lacks the relational inductive bias essential for modeling structured data like WSIs, where tile relationships are not sequential but spatial and high-dimensional~\cite{yu2024mambaout,ren2024can}. This tension raises a central question: How can we preserve the powerful relational or sequential modeling capabilities of GNNs and Transformers while achieving the efficiency of Mamba?
% How can we achieve an optimal balance between modeling capacity and computational complexity?

To this end, we propose the \textbf{WSI-HGMamba}, a novel framework that unifies the stronger relational modeling capacity of Hypergraph Neural Networks with the linear-time sequential modeling efficiency of Mamba. 
% framework, which addresses the challenges of large-scale WSI analysis by combining the best of both aspects, \textit{i.e.}, powerful relational modeling and highly efficient sequence learning. 
% Central to our approach is the proposed HGMamba block, which integrates the high-order relational modeling strengths of HGNNs~\cite{gao2024hypergraph,cheng2024multi,lu2023exploring} with the computational efficiency of sequence learning models like Mamba~\cite{xing2024segmamba,lin2024event,liu2025hydra}.
% The HGMamba block consists of three key modules, including Message Passing, hypergraph Scanning \& Flattening, and feature aggregation using the Bidirectional State Space Model (Bi-SSM). 
% This design enables the model to achieve Transformer-level performance with HGNN or Mamba-level computational costs, achieving up to a $7 \times$ reduction in FLOPs compared to Transformer and Graph Transformer-based models. 
In the proposed approach, WSIs are partitioned into tiles, and a pre-trained tile encoder extracts tile-level features. The hypergraph construction is based on both spatial-driven (rule-based) and data-driven (similarity-based) adjacency schemes. Inside each HGMamba block, we introduce the Hypergraph Scanning \& Flatten mechanism that converts high-order relational structures into ordered sequences amenable to Mamba-based processing. The message-passed features are then dynamically updated through the Bi-SSM, allowing the model to preserve the relational context while maintaining computational efficiency. Compared with Transformer and Graph Transformer baselines, WSI-HGMamba achieves up to $7\times$ FLOPs reduction while maintaining comparable or superior classification accuracy on the histopathology benchmarks.
% We construct a WSI hypergraph by integrating both rule-based adjacency and similarity-based adjacency. 
% The resulting hypergraph is then processed by the proposed HGMamba network, which consists of multiple HGMamba blocks that perform key operations: Message Passing, Hypergraph Scanning \& Flattening, and feature aggregation using the Bidirectional State Space Model (Bi-SSM). These blocks effectively combine the robust relational modeling capacity of HGNNs with the low-complexity sequence learning capabilities of Mamba, enabling the model to achieve Transformer-level performance with Mamba-level computational costs, reducing FLOPs by up to $7 \times$ compared to Transformer and Graph Transformer-based models.
% Our experimental results demonstrate that the WSI-GMamba framework outperforms traditional methods while maintaining low memory usage and computational overhead. 
% WSI-GMamba represents an advancement in MIL analysis, positioning it as a potential next-generation backbone for WSI understanding.
The main contributions are as follows:

\begin{itemize}
    \item We propose \textbf{WSI-HGMamba}, a novel framework that integrates Hypergraph Neural Networks with Mamba-based State Space Models for efficient and expressive WSI analysis.
    \item We design the HGMamba block, combining message passing, hypergraph scanning \& flattening, and bidirectional state space modeling on a hypergraph constructed from both spatial-driven and data-driven adjacency.
    \item Our framework achieves Transformer-level performance with up to $7\times$ lower FLOPs, enabling scalable slide-level understanding on large-scale WSIs.
\end{itemize}

\section{Related Work}
\subsubsection{Relational Modeling for MIL.}
In the domain of tissue pathology, Multiple Instance Learning (MIL) has increasingly relied on graph-based representations that treat image tiles as distinct entities. 
For example, Zhao \textit{et al.}~\cite{zhao2020predicting} introduce a multiple instance learning framework with deep graph convolution networks to predict lymph node metastasis from histopathological images. Chen \textit{et al.}~\cite{chen2021whole} utilize GCN to capture both local and global topological structures for survival prediction. Chan \textit{et al.}~\cite{chan2023histopathology} focus on the diversity within the tissue microenvironment, creating heterogeneous graphs to represent WSIs, incorporating the complex relationships between various types of nuclei. Yu \textit{et al.}~\cite{yu2024dualgcn} propose DualGCN-MIL, a WSI classification model that leverages dual relationship graph learning to enhance representation learning in weakly supervised settings. 

Recent advances in histopathological MIL have moved beyond pairwise graph models to hypergraph neural networks, which capture higher-order, non-pairwise relationships among WSI patches. For example, Di \textit{et al.}~\cite{di2022big} introduce b‑HGFN, a Big‑Hypergraph Factorization Neural Network that embeds large-scale vertices and hyperedges into low-dimensional semantic spaces for efficient survival prediction on gigapixel WSIs. In a related study, Di \textit{et al.}~\cite{di2022generating} propose a method to generate high-order hypergraph representations of WSIs by constructing instance-level hyperedges based on spatial proximity and phenotypic similarity of image patches. Shi \textit{et al.}~\cite{shi2024masked} propose Masked Hypergraph Learning, leveraging hypergraph convolution under weak supervision to improve slide-level classification in MIL settings. Additionally, Han \textit{et al.}~\cite{han2025inter} introduce inter-intra hypergraph computation, a novel dual-level hypergraph model that jointly encodes intra-slide and inter-slide relationships. 
% In tissue pathology, Multiple Instance Learning (MIL) increasingly employs graph-based representations of image tiles. Zhao \textit{et al.}~\cite{zhao2020predicting} use deep graph convolution networks for lymph node metastasis prediction, while Chen \textit{et al.}~\cite{chen2021whole} leverage GCNs to model local and global structures for survival prediction. Chan \textit{et al.}~\cite{chan2023histopathology} construct heterogeneous graphs to capture tissue microenvironment diversity, and Yu \textit{et al.}~\cite{yu2024dualgcn} propose DualGCN-MIL, enhancing WSI classification via dual relationship graph learning. 
% Graph Transformers further advance representation learning. Zheng \textit{et al.}~\cite{zheng2022graph} integrate them for disease severity prediction, Shirzad \textit{et al.}~\cite{shirzad2023exphormer} introduce Expformer for efficient graph-structured data processing, and Shi \textit{et al.}~\cite{shi2024integrative} develop an Integrative Graph-Transformer (IGT) framework to refine WSI analysis.

\begin{figure*}[t]
\includegraphics[width=\textwidth]{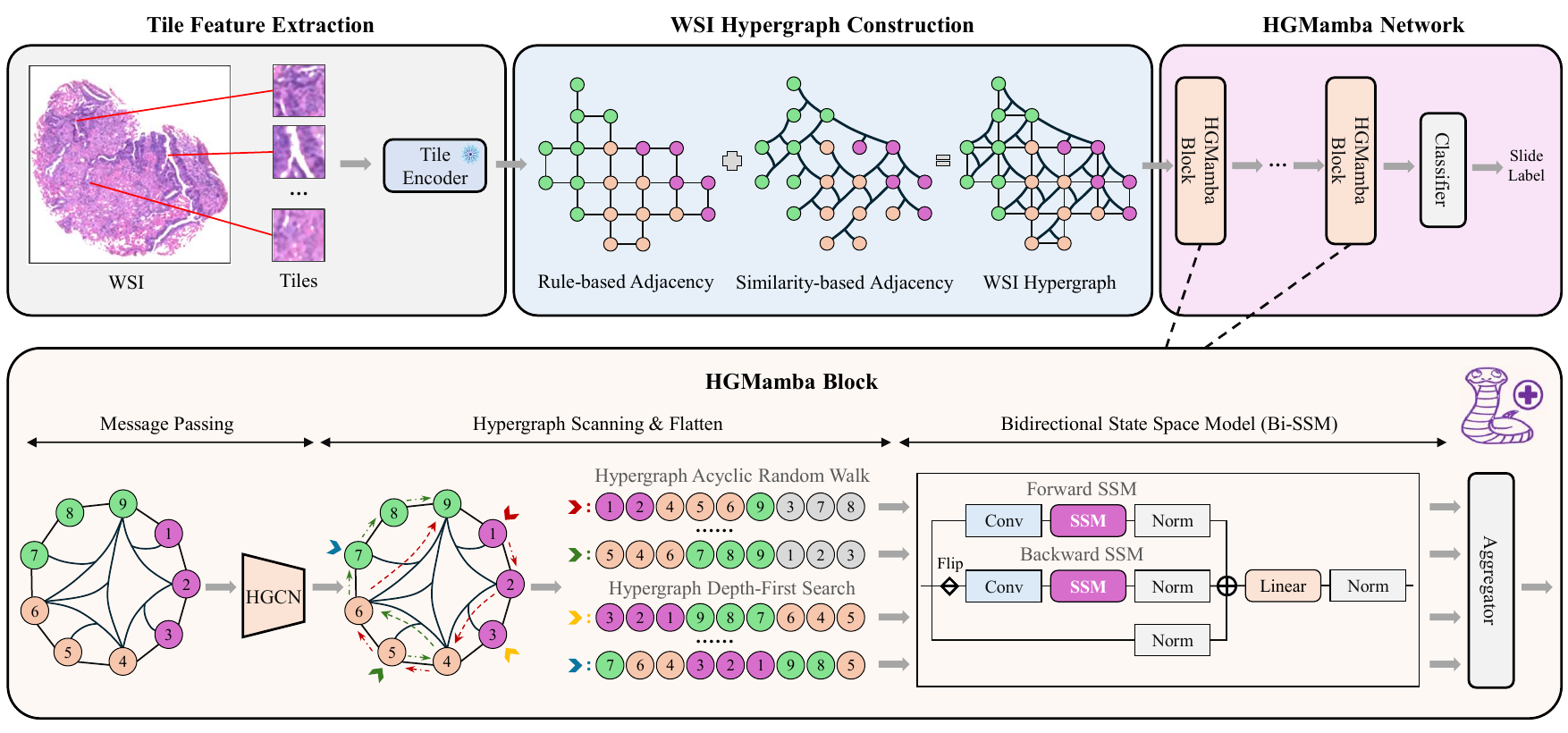}
\caption{The overall pipeline for WSI-HGMamba framework. WSIs are partitioned into tiles, and a pre-trained tile encoder is employed to extract tile-level features. The WSI hypergraph is constructed by integrating the rule-based adjacency and the similarity-based adjacency. This hypergraph is then processed by the proposed HGMamba network, which consists of multiple HGMamba blocks. Each block performs three key operations: message passing, hypergraph scanning \& flatten, and feature aggregation using the Bi-SSM. Finally, a classifier is applied to predict the slide-level labels.} \label{pipeline}
\end{figure*}

\subsubsection{Sequential Modeling for MIL.}
% Sequence modeling techniques have also been explored for analyzing WSIs, particularly leveraging transformer architectures. 
Sequential modeling techniques, particularly transformer architectures, have been explored for WSI analysis~\cite{chen2021multimodal,wang2022transformer,li2023vision,li2024rethinking,gao2024transformer,yan2025pathorchestra}. Chen \textit{et al.}~\cite{chen2021multimodal} propose the Multimodal Co-Attention Transformer (MCAT) to integrate histopathological and clinical data for survival prediction. Shao \textit{et al.}~\cite{shao2021transmil} develop TransMIL, leveraging transformers to capture both morphological and spatial information in WSIs. 

Additionally, several works introduce Graph Transformers to enhance the learning capability. Zheng \textit{et al.}~\cite{zheng2022graph} integrate Graph Transformers to enhance the spatial representation of WSIs, enabling the prediction of disease severity. Shirzad \textit{et al.}~\cite{shirzad2023exphormer} develop Expformer, a sparse transformer model designed to handle graph-structured data efficiently, which can be applied to histopathological analysis. Shi \textit{et al.}~\cite{shi2024integrative} propose an Integrative Graph-Transformer (IGT) framework for WSI representation and classification, combining graph-based and transformer-based approaches to improve histopathology image analysis.

Despite their ability to capture long-range dependencies, transformer-based and graph transformer-based methods face high computational complexity with gigapixel WSIs. Li \textit{et al.}~\cite{li2024rethinking} introduce LongMIL to optimize transformer efficiency for long-context WSI analysis. To further enhance efficiency, Yang \textit{et al.}~\cite{yang2024mambamil} propose MambaMIL, leveraging State Space Models (SSMs) for long-sequence modeling. While Mamba-based methods reduce computational costs, they may still fall short of transformers or GNNs in capturing complex WSI representations. How to preserve Transformer-level performance while maintaining Mamba-level efficiency for WSI understanding is an important avenue for exploration.

\section{Methodology}
In this section, we describe the proposed WSI-HGMamba framework for whole-slide image (WSI) analysis. As illustrated in Fig.~\ref{pipeline}, the framework consists of four main stages: (1) partitioning the WSI into tiles and extracting tile-level features; (2) constructing a WSI-level hypergraph by combining Rule-based Adjacency with Similarity-based Adjacency; (3) processing the hypergraph via the proposed HGMamba network, which stacks multiple HGMamba blocks; and (4) aggregating the resulting representations via a classifier for slide-level prediction. 
% The subsequent subsections detail each step of the method.

\subsection{Tile Feature Extraction}
Given a whole-slide image, we define it as a collection of tiles:
\begin{equation} 
\mathcal{W} = \{\,T_1,\ T_2,\ \dots,\ T_N\},
\end{equation}
where each $T_i$ corresponds to a small image patch (\textit{i.e.}, tile) obtained by a fixed-size sliding window (\textit{e.g.}, $512 \times 512$ pixels). We feed each tile $T_i$ into a pre-trained tile encoder (\textit{e.g.}, a ResNet or ViT) to obtain a feature vector $\mathbf{x}_i \in \mathbb{R}^d$, where $d$ is the feature dimension. Stacking the features of all $N$ tiles yields the feature matrix:
% \begin{equation}
% \mathbf{X} = \begin{bmatrix}
% \mathbf{x}_1^\top \\
% \mathbf{x}_2^\top \\
% \vdots \\
% \mathbf{x}_N^\top
% \end{bmatrix}
% \in \mathbb{R}^{N \times d}.
% \end{equation}
\begin{equation}
\mathbf{X} = 
\begin{bmatrix} 
\mathbf{x}_1, & \mathbf{x}_2, & \dots, & \mathbf{x}_N 
\end{bmatrix} 
\in \mathbb{R}^{N \times d}.
\end{equation}
These tile-level features form the basis for our subsequent hypergraph construction and analysis.

\subsection{WSI Hypergraph Construction}
To capture both in-context adjacency (\textit{e.g.}, top / bottom / left / right neighbors) and long-range semantic similarity, we construct two types of adjacencies, spatial-driven (rule-based) and data-driven (similarity-based), and then integrate them into a unified WSI hypergraph. 

\subsubsection{Rule-based Adjacency.}
We introduce standard edges between tiles that are spatially directly adjacent in the WSI layout, \textit{i.e.}, those sharing boundaries on the top, bottom, left, or right. Letting $\mathcal{N}(i)$ to denote the set of neighbors of tile $i$ under these adjacency rules, the rule-based edge connectivity can be represented using an incidence matrix $\mathbf{H}^{\text{rule}} \in \{0,1\}^{N \times M}$, which can be considered a degenerate hypergraph structure where each hyperedge connects exactly two nodes:
\begin{equation}
\mathbf{H}^{\text{rule}}_{ie}  
=  
\begin{cases}
1, & \text{if } e \text{ connects } i \text{ and } j \text{ where } j \in \mathcal{N}(i),\\
0, & \text{otherwise}.
\end{cases}
\end{equation}
This local connectivity is critical for preserving neighborhood context in pathological images.

\subsubsection{Similarity-based Adjacency.}
Local adjacency alone is often insufficient to capture meaningful long-range semantic relationships. 
Hence, we construct similarity-based hyperedges to connect tiles exhibiting high semantic similarity. Specifically, for each tile $i$, we identify its top-$K$ most similar tiles $\mathcal{N}_K(i)$ based on cosine similarity, forming a hyperedge that directly connects these tiles. Formally, the similarity-based hypergraph can be represented using an incidence matrix $\mathbf{H}^{\text{sim}} \in \{0,1\}^{N \times M'}$ as follows:
% Hence, we also compute a similarity metric $\mathrm{Sim}(\mathbf{x}_i, \mathbf{x}_j)$ for each pair $(i,j)$. For each tile $i$, we identify the top-$K$ most similar tiles $\mathcal{N}_K(i)$ and connect them together via a hyperedge. Formally, the similarity-based incidence matrix $\mathbf{A}^{\text{sim}} \in \{0,1\}^{N \times M}$ is defined as
\begin{equation}
\mathbf{H}^{\text{sim}}_{ie} = 
\begin{cases}
1, & \text{if } e \text{ connects } i \text{ and } j \text{ where } j \in \mathcal{N}_k(i),\\
0, & \text{otherwise},
\end{cases}
\end{equation}
where the semantic similarity between tile $i$ and $j$ is computed by the cosine similarity measure:
\begin{equation}
\mathrm{Sim}(\mathbf{x}_i, \mathbf{x}_j) = \frac{\mathbf{x}_i \cdot \mathbf{x}_j}{|\mathbf{x}_i||\mathbf{x}_j|}.
\end{equation}
This hypergraph structure explicitly connects distant yet semantically related tiles, enabling hypergraph neural network operations to directly leverage these "long-range" semantic interactions.

% If it exceeds a threshold $\theta$, an edge is introduced to indicate high semantic affinity. Formally, the similarity-based adjacency matrix $\mathbf{A}^{\text{sim}} \in \{0,1\}^{N \times N}$ is given by
% \begin{equation}
% \mathbf{A}^{\text{sim}}_{ij}  
% =  
% \begin{cases}
% 1, & \text{if } \mathrm{Sim}(\mathbf{x}_i, \mathbf{x}_j) \ge \theta,\\
% 0, & \text{otherwise},
% \end{cases}
% \end{equation}

\subsubsection{Unified WSI Hypergraph.}
The final WSI hypergraph incidence matrix $\mathbf{H}$ results from combining the rule-based and similarity-based incidence matrices:
% \begin{equation}
% \mathbf{H} = \mathbf{H}^{\text{rule}} + \mathbf{H}^{\text{sim}},
% \end{equation}
\begin{equation}
\mathbf{H} = [\mathbf{H}^{\text{rule}}, \mathbf{H}^{\text{sim}}],
\end{equation}
where $[\cdot,\cdot]$ denotes horizontal concatenation. Together, $\mathcal{G} = \{\mathbf{X}, \mathbf{H}\}$ defines our unified WSI hypergraph, where $\mathbf{X}$ represents node features, and $\mathbf{H}$ captures both in-context adjacency and long-range semantic connectivity via hyperedges.

\subsection{HGMamba Block}
An HGMamba block receives node features $\mathbf{X} \in \mathbb{R}^{N \times d}$ and the hypergraph incidence matrix $\mathbf{H}$ as inputs. After processing, it outputs enhanced node features $\mathbf{X}^{(l)}$ to either subsequent blocks or the final classifier.

\subsubsection{Message Passing.} 
We update node features through hypergraph convolution~\cite{feng2019hypergraph}, capturing high-order interactions via:
\begin{equation}
\mathbf{X}^{(l)} = 
\sigma \left( 
\mathbf{D}_v^{-\frac{1}{2}} \, 
\mathbf{H} \, 
\mathbf{W}_e \, 
\mathbf{D}_e^{-1} \, 
\mathbf{H}^\top \, 
\mathbf{D}_v^{-\frac{1}{2}} \, 
\mathbf{X} \, 
\mathbf{W}^{(l-1)} 
\right),
\end{equation}
where $\mathbf{D}_v \in \mathbb{R}^{N \times N}$ and $\mathbf{D}_e \in \mathbb{R}^{E \times E}$ arepresent diagonal matrices of node and hyperedge degrees, respectively. $\mathbf{W}_e \in \mathbb{R}^{E \times E}$ and $\mathbf{W}^{(l-1) \in \mathbb{R}^{d \times d}}$ are learnable parameter matrixs, and $\sigma(\cdot)$ denotes the ReLU activation function.

% \textbf{Message Passing.} 
% We first update $\mathbf{X}^{(0)}$ by aggregating neighbors’ features, for instance, via a common Graph Convolutional Network (GCN) layer:
% \begin{equation}
% \mathbf{X}^{(l)}  
% =  
% \sigma \Big(  
% \tilde{\mathbf{D}}^{-\frac{1}{2}}  
% \,\tilde{\mathbf{A}}\,  
% \tilde{\mathbf{D}}^{-\frac{1}{2}}  
% \,\mathbf{X}\,\mathbf{W}^{(l-1)}  
% \Big),
% \end{equation}

% where $\tilde{\mathbf{A}} = \mathbf{A} + \mathbf{I}$, $\tilde{\mathbf{D}}$ is the degree matrix of $\tilde{\mathbf{A}}$, $\mathbf{W}^{(0)}$ is a learnable parameter matrix, and $\sigma(\cdot)$ is the ReLU activation. 

\subsubsection{Hypergraph Scanning \& Flattening.} 
To exploit the sequential modeling capabilities of the State Space Model, we convert the hypergraph into a set of sequences by traversing hyperedges explicitly. Given a set of sequences $\mathcal{S} = \{S^{(m)}\}_{m=1}^{M}$, each sequence is generated by the two proposed hypergraph traversal strategies:

- Hypergraph Depth-First Search (H-DFS): Starting from a randomly chosen root node $r$, traversal proceeds by selecting nodes within connected hyperedges. When entering a hyperedge, traversal may continue to any previously unvisited node within the hyperedge, thereby effectively capturing high-order node connectivity.
% from a randomly chosen root node $r$, we traverse the graph until all $N$ nodes are visited. 

- Acyclic Random Walk (H-ARW):
we start at $r$ and randomly select subsequent neighbors within the hyperedge without revisiting any node, thereby constructing a path of fixed length $T$. If the walk terminates before reaching length $T$, we apply padding for the remaining $N-T$ nodes.

Each hypergraph traversal yields a sequence of node features:
\begin{equation}
S^{(m)}  
= \bigl(\mathbf{s}^{(m)}_1,\ \mathbf{s}^{(m)}_2,\ \dots,\ \mathbf{s}^{(m)}_N\bigr),
\end{equation}
where $\mathbf{s}^{(m)}_t \in \mathbb{R}^d$ denotes the feature of the node visited at step $t$, and $N$ is a maximum sequence length.  
H-DFS captures structural connectivity with varying node orders from the same root, while H-ARW allows stochastic exploration of arbitrary regions of the hypergraph, potentially capturing diverse connectivity.. Combining these scanning methods benefits the SSM model by exposing it to a broad range of node sequences.

\subsubsection{Bidirectional State Space Model.}
% The introduced Bidirectional State-Space Model (Bi-SSM) processes the input sequence by leveraging both forward and backward components, with each component consisting of a sequence of operations: a 1D convolution, a State-Space Model (SSM), and normalization. The outputs from both directions are then combined through a residual addition and passed through a linear transformation followed by normalization, as shown in Fig.~\ref{pipeline}. 
Given an input sequence $S^{(m)} = (\mathbf{s}^{(m)}_1, \mathbf{s}^{(m)}_2, \dots, \mathbf{s}^{(m)}_T)$, where each $\mathbf{s}^{(m)}_t \in \mathbb{R}^d$ represents the feature of the $t^{th}$ token, we process it bidirectionally. In the forward direction, we first apply a 1D convolution to the input sequence $\mathbf{z}^{f,(m)}_t = \text{Conv}(\mathbf{s}^{(m)}_t)$. The result is then passed through the SSM~\cite{mamba,mamba2} module, which captures the sequential dependencies $\mathbf{z}^{f,(m)}_t = \text{SSM}(\mathbf{z}^{f,(m)}_t)$. Afterward, we apply normalization $\mathbf{z}^{f,(m)}_t = \text{Norm}(\mathbf{z}^{f,(m)}_t)$. The same operations are applied in the backward pass, yielding $\mathbf{z}^{b,(m)}_t$. 
The outputs are merged via residual addition and passed through a linear and norm layer.
% The outputs are then combined via residual addition. This combined result is passed through a linear layer and a norm layer.

% Similarly, in the backward direction, we apply the same operations. We utilize a 1D convolution $\mathbf{z}^{b,(m)}_t = \text{Conv}(\mathbf{s}^{(m)}_t)$, followed by the SSM operation $\mathbf{z}^{b,(m)}_t = \text{SSM}(\mathbf{z}^{b,(m)}_t)$ and then normalization $\mathbf{z}^{b,(m)}_t = \text{Norm}(\mathbf{z}^{b,(m)}_t)$. The outputs from the forward and backward blocks are then combined via residual addition. This combined result is passed through a linear layer and a norm layer.

Once the Bi-SSM block has processed the sequences and produced the sequence representations $\{\tilde{\mathbf{Z}}^{(1)}, \dots, \tilde{\mathbf{Z}}^{(M)}\}$, we apply an aggregator to combine these representations, whose purpose is to aggregate token information corresponding to the same node, mapping the $m$-dimensional sequence back to the original hypergraph structure. 
% For example, if a node appears at different positions in different sequences, its corresponding token representations are grouped and aggregated accordingly. 
% average the corresponding tokens across the $M$ sequences, 
Specifically, for each node $t$, we compute its representation by averaging its corresponding tokens across all sequences where it appears:
% \begin{equation}
%      \tilde{\mathbf{z}}_t = \frac{1}{M} \sum_{m=1}^{M}  \tilde{\mathbf{z}}^{(m)}_t.
% \end{equation}
\begin{equation}
\tilde{\mathbf{z}}_t = \frac{1}{|S_t|} \sum_{m\in S_t} \tilde{\mathbf{z}}^{(m)}_t,
\end{equation}
where $S_t$ denotes the set of sequences in which node $t$ appears.
After this aggregation, the resulting sequence of node embeddings can be reinterpreted as a hypergraph and processed by the next HGMamba block. Finally, we pass $\tilde{\mathbf{Z}}=\{\tilde{\mathbf{z}}_1, \tilde{\mathbf{z}}_2,\dots,\tilde{\mathbf{z}}_N\}$ into the ABMIL~\cite{ilse2018attention} classifier to obtain the predicted slide-level label, with cross-entropy loss used for training.

\begin{table*}[t]
\renewcommand{\arraystretch}{1.3} 
\centering
\caption{Performance comparison of GNN, Mamba, Transformer, and Graph Transformer-based approaches. The proposed WSI-HGMamba achieves superior performance to Transformer and Graph Transformer-based methods with similar GFLOPs and memory usage of GNN-based methods.}
\resizebox{\textwidth}{!}{
\begin{tabular}{l|l|c|c|c|c|c|c|c|c|c|c}
\hline
\multirow{2}{*}{\textbf{Type}} & \multirow{2}{*}{\textbf{Model}} & \multicolumn{2}{c|}{\textbf{TCGA-ESCA}} & \multicolumn{2}{c|}{\textbf{TCGA-NSCLC}} & \multicolumn{2}{c|}{\textbf{TCGA-RCC}} & \multicolumn{2}{c|}{\textbf{Prost}} & \textbf{FLOPs} & \textbf{\footnotesize Mem.} \\
\cline{3-10}
 & & \textbf{AUC} & \textbf{ACC} & \textbf{AUC} & \textbf{ACC} & \textbf{AUC} & \textbf{ACC} & \textbf{AUC} & \textbf{F1} & \textbf{(G)} & \textbf{(GB)} \\
\hline
\textbf{GNN} & GAT           & 85.8  & 86.4  & 89.2  & 86.2  & 92.1  & 88.4  & 92.3  & 52.4  & 1.4  & 0.2 \\
             & GCN           & 90.8  & 90.9  & 90.1  & 86.0  & 93.0  & 89.1  & 93.2  & 54.6  & 0.8  & 0.2 \\
             & GIN           & 91.6  & 90.9  & 90.2  & 87.1  & 93.5  & 90.0  & 93.5  & 59.7  & 1.0    & 0.2 \\
             & GCN-MIL (2020)~\cite{zhao2020predicting} & 87.1  & 88.0    & 92.4  & 88.1    & 94.3  & 88.4  & 93.4  & 56.7  & 1.2  & 0.3  \\
             & Patch-GCN (2021)~\cite{chen2021whole} & 91.3  & 91.1  & 95.0  & 88.8    & 95.1  & 89.7  & 94.1  & 64.3  & 1.5  & 0.4  \\
         & HEAT (2023)~\cite{chan2023histopathology}    & 92.8  & 92.2  & 94.3  & 87.7  & 94.1  & 90.2  & 93.6  & 62.3  & 1.3  & 0.3  \\
             & DualGCN-MIL (2024)~\cite{yu2024dualgcn} & 91.7  & 92.5  & 94.1  & 89.5  & 95.2  & 91.1  & 95.5  & 74.2  & 1.6  & 0.5  \\
\hline
\textbf{Mamba} & MambaMIL (2024)~\cite{yang2024mambamil} & 91.3  & 92.2  & 91.2  & 88.2  & 94.2  & 90.2  & 94.2  & 70.4  & 1.9  & 0.9  \\
 & 2DMamba (2025)~\cite{zhang20252dmamba} & 92.4  &  93.0 & 93.5  & 88.0  & 95.1  & 90.0  & 95.1  & 74.4  & 1.8  & 0.7  \\
\hline
\textbf{Trans.} & MCAT (2021)~\cite{chen2021multimodal}    & 94.5  & 93.2  & 94.1  & 90.1  & 95.6  & 91.0  & 96.2  & 78.1  & 8.7  & 1.4  \\
                  & LongMIL (2024)~\cite{li2024rethinking} & 95.8  & 93.0  & 96.0  & 92.3    & 97.8    & 92.1  & 96.1  & 79.4  & 10.3  & 2.1  \\
\hline
\textbf{GTrans.} & GTP (2022)~\cite{zheng2022graph} & 95.3  & 94.0  & 95.8  & 90.5  & 97.7  & 91.4  & 95.2  & 78.5  & 9.5  & 1.7  \\
                    & Expformer (2023)~\cite{shirzad2023exphormer} & 96.4  & 93.5  & 94.9  & 91.1  & 96.8  & 91.2  & 95.7  & 79.1  & 7.5  & 1.4  \\
                     & IGT (2024)~\cite{shi2024integrative}       & 96.1  & 93.7  & 96.7  & 91.6  & 98.4  & 92.4  & 96.3  & 80.3  & 12.5  & 2.5  \\
\hline
\textbf{Ours}  & \textbf{WSI-GMamba}       & 97.7  & 94.2  & 96.9  & 91.6  & 98.0    & 92.4  & 97.1  & \textbf{82.4}  & 1.8  & 0.8  \\
   & \textbf{WSI-HGMamba}       & \textbf{98.7}  & \textbf{95.8}  & \textbf{98.4}  & \textbf{92.9}  & \textbf{99.3}    & \textbf{93.6}  & \textbf{98.8}  & 82.2  & 2.2  & 1.0  \\
\hline
\end{tabular}
}
\label{tab:sota_comparison}
\end{table*}

\section{Experiments and Results}
\subsection{Experimental Settings}
\subsubsection{Datasets.} 
We conduct experiments on three widely used public datasets, TCGA-ESCA, TCGA-NSCLC, and TCGA-RCC, as well as the in-house dataset Prost. 
TCGA-ESCA focuses on esophageal carcinoma and includes diagnostic WSIs with slide-level labels for histologic subtypes. 
TCGA-NSCLC is composed of two sub-cohorts from lung cancer: lung adenocarcinoma (LUAD) and lung squamous cell carcinoma (LUSC). It covers a spectrum of morphological variability, offering a challenging setting for subtype discrimination.
TCGA-RCC includes renal cell carcinoma samples, comprising clear cell, papillary, and chromophobe subtypes. 
% All public datasets are sourced from The Cancer Genome Atlas (TCGA) and preprocessed using standardized pipelines including tissue detection, tiling at $20\times$ magnification, and removal of background or low-information tiles.
Prost is an in-house dataset from an anonymous hospital, specifically focused on prostate Gleason grading, comprising 1,042 WSIs categorized into four classes: negative, grade 3, grade 4, and grade 5. All datasets are split into training, validation, and test sets using a ratio of 7:2:1.

\subsubsection{Implementations.}
We divide the WSI into non-overlapping tiles at a resolution of $512\times 512$ and extract embeddings using ResNet50~\cite{he2016deep}. 
% A similarity-based adjacency graph is constructed by connecting vertices with cosine similarity $\theta >0.9$.
In the construction of the similarity-based adjacency, the value of $K$ for selecting the top-$K$ neighboring tiles is set to 3.
For Hypergraph Scanning, we generate $M=8$ traversal sequences using H-DFS and H-ARW, with the latter having a fixed length $T=0.7N$. The framework comprises $2$ HGMamba layers, optimized using Adam~\cite{kingma2014adam}, with the learning rate $0.001$ and weight decay 0.0005 using MultiStepLR scheduler. All experiments are conducted for 120 epochs with a batch size of 12 on an NVIDIA RTX 4090 GPU.

\subsubsection{Baselines.}
We design a series of baselines and model variants to evaluate the performance of WSI classification. First, we adopt GCN to model spatial dependencies between tiles based on a predefined adjacency graph, and Bi-SSM to capture long-range contextual dependencies efficiently in a bidirectional manner. To combine spatial and sequential modeling, we introduce WSI-GMamba, which integrates GCN with Bi-SSM. Furthermore, we improve spatial modeling by upgrading GCN to HGCN, resulting in WSI-HGMamba, which jointly captures high-order spatial relations and global contextual patterns.

\begin{figure*}[t]
\includegraphics[width=\textwidth]{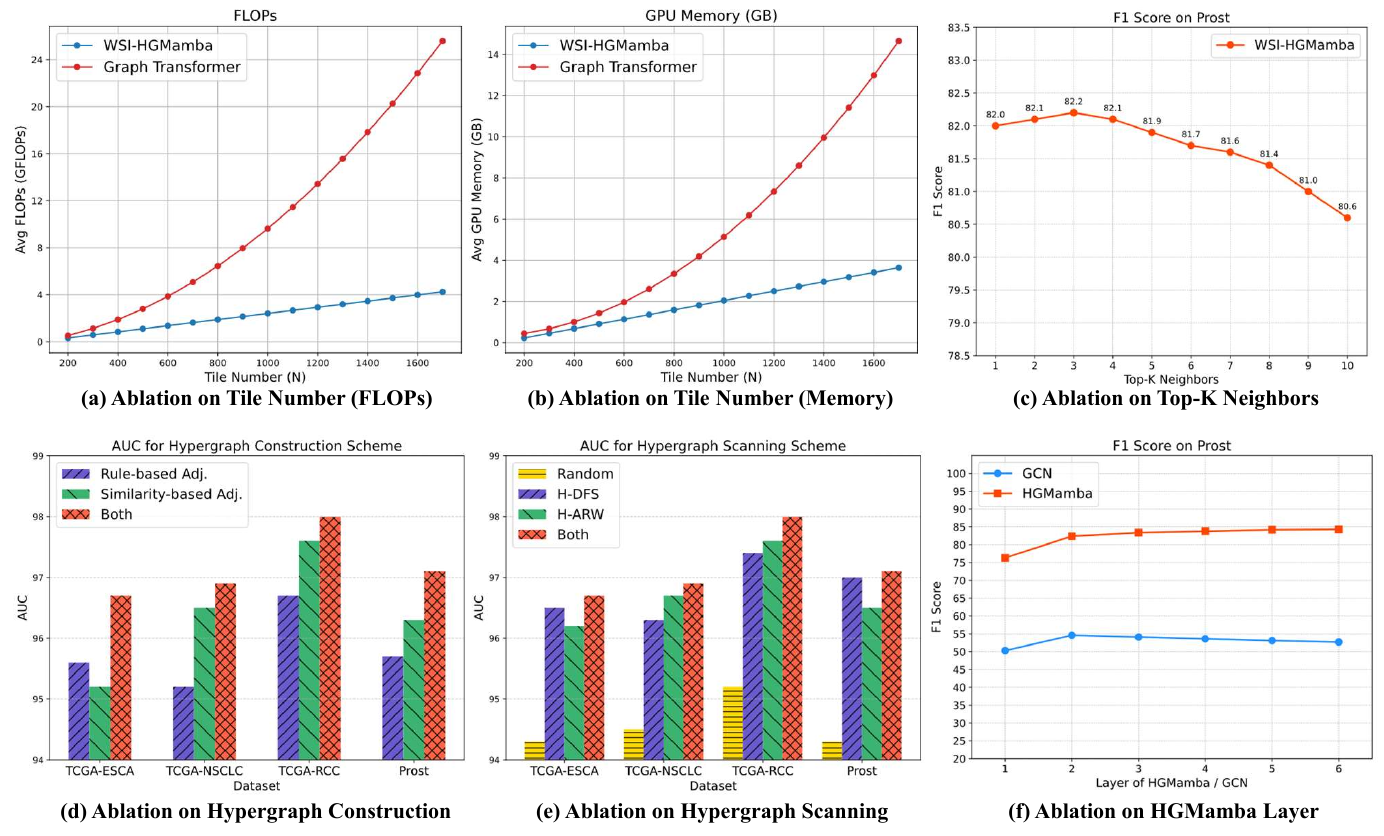}
\caption{Ablation experiments on the proposed WSI-HGMamba. (a) and (b) compare the computational costs (FLOPs) and GPU memory usage of WSI-HGMamba and Graph Transformer as tile numbers increase. (c) shows the F1 scores on the Prost dataset for different values of Top-$K$ neighbors in the hypergraph construction. (d) compares the AUC scores of various hypergraph construction schemes (Rule-based, Similarity-based, and Both) across datasets. (e) presents AUC comparison of hypergraph scanning methods (Random, H-DFS, H-ARW, and Both). (f) shows the F1 scores for varying HGMamba and GCN layer numbers on the Prost dataset.} \label{ablation_experiments}
\end{figure*}

\subsection{Experimental Results}
\subsubsection{Comparison with SOTAs.} 
To comprehensively evaluate the effectiveness of the proposed models, we compare them against a broad set of SOTA methods across four categories: GNN-based, Mamba-based, Transformer-based, and Graph Transformer-based methods. GNN models such as GAT, GCN, GIN, and Patch-GCN~\cite{chen2021whole} focus on pairwise spatial modeling of WSI tiles, while recent works like HEAT~\cite{chan2023histopathology} and DualGCN-MIL~\cite{yu2024dualgcn} incorporate refined graph structures for improved performance. Mamba-based methods, including MambaMIL~\cite{yang2024mambamil} and 2DMamba~\cite{zhang20252dmamba}, leverage SSMs to model long-range dependencies efficiently without attention mechanisms. Transformer-based approaches like MCAT~\cite{chen2021multimodal} and LongMIL~\cite{li2024rethinking} utilize global attention to capture contextual information across tiles, whereas Graph Transformer variants such as GTP~\cite{zheng2022graph}, Expformer~\cite{shirzad2023exphormer}, and IGT~\cite{shi2024integrative} combine graph structure priors with attention to model complex interactions.

As shown in Table~\ref{tab:sota_comparison}, the proposed models WSI-GMamba and WSI-HGMamba consistently achieve superior performance across all benchmarks. WSI-GMamba, which combines GCN with Bi-SSM, already surpasses most baselines, while WSI-HGMamba, enhanced with hypergraph modeling, achieves the best overall results. Specifically, WSI-HGMamba achieves AUC scores of 98.7\% on TCGA-ESCA, 98.4\% on TCGA-NSCLC, 99.3\% on TCGA-RCC, and 98.8\% on Prost, significantly outperforming GNN-based and Mamba-based methods, and even surpassing state-of-the-art Transformer and Graph Transformer approaches, highlighting its ability to capture both relational structures and sequence dependencies. 
In particular, the performance is achieved with considerably lower computational cost, only 2.2 GFLOPs and 1.0 GB of memory, compared to heavier models like IGT~\cite{shi2024integrative}, which require 12.5 GFLOPs and 2.5 GB. These results highlight the effectiveness of combining efficient SSM-based sequential modeling with spatial and high-order structural modeling, enabling our approach to achieve both state-of-the-art performance and practical scalability.
% As shown in Table~\ref{tab:sota_comparison}, WSI-HGMamba achieves SOTA performance with lower computational cost compared to GNN, Mamba, Transformer, and Graph Transformer models. Against GNN-based and Mamba-based methods, WSI-HGMamba delivers notably higher AUC, ACC, and F1 (e.g., 96.7\% AUC vs. 91.7\% for DualGCN-MIL~\cite{yu2024dualgcn}) while maintaining similar GFLOPs and memory usage. This highlights its ability to capture both relational structures and sequence dependencies.
% Compared to Transformer or Graph Transformer models, WSI-HGMamba achieves comparable or superior performance while reducing FLOPs by up to $7\times$. 
% % For example, IGT (2024) achieves 96.8% AUC with 7.5 GFLOPs, whereas WSI-HGMamba reaches 96.7% AUC with only 1.8 GFLOPs. Similarly, LongMIL (2024) requires 8.7 GFLOPs but performs worse than WSI-HGMamba.
% WSI-HGMamba balances efficiency and accuracy, surpassing or matching SOTA methods while maintaining the lightweight nature of GNNs and Mamba.

\begin{table*}[t]
\centering
\caption{Performance comparison (\%) for HGMamba block on four benchmarks.}
\label{tab:performance_comparison}
\begin{tabular}{lcccc}
\toprule
\textbf{Method} & \textbf{TCGA-ESCA (ACC)} & \textbf{TCGA-NSCLC (ACC)} & \textbf{TCGA-RCC (ACC)} & \textbf{Prost (F1)} \\
\midrule
GCN (Baseline)    & 90.9 & 86.0 & 89.1 & 54.6 \\
SSM (Baseline)    & 88.6 & 87.4 & 89.0 & 65.3 \\
Bi-SSM            & 89.5 & 87.6 & 89.5 & 69.3 \\
GCN + SSM         & 92.1 & 90.2 & 90.2 & 75.3 \\
GCN + Bi-SSM      & 92.4 & 91.2 & 90.7 & 78.4 \\
\hline
\textbf{WSI-GMamba}        & 94.2 & 91.6 & 92.4 & \textbf{82.4} \\
\textbf{WSI-HGMamba}       & \textbf{95.8} & \textbf{92.9} & \textbf{93.6} & 82.2 \\
\bottomrule
\end{tabular}
\end{table*}

\subsubsection{Ablation on HGMamba Block.}
We conduct ablation studies to evaluate the effect of different modeling components within the HGMamba block across four benchmarks. As shown in Table~\ref{tab:performance_comparison}, both GCN and SSM individually serve as baselines. Replacing the vanilla SSM with a bidirectional version (Bi-SSM) consistently improves performance across all datasets, confirming the benefit of bidirectional context modeling. Combining GCN with SSM-based methods further boosts accuracy. GCN + SSM achieves moderate gains, while GCN + Bi-SSM delivers substantial improvements, demonstrating the complementarity of relational and sequential modeling. The proposed WSI-GMamba outperforms all baselines and intermediate variants. Furthermore, WSI-HGMamba, which incorporates hypergraph-based high-order spatial modeling instead of pairwise GCN, achieves the best performance across the board: 95.8\% on TCGA-ESCA, 92.9\% on TCGA-NSCLC, 93.6\% on TCGA-RCC, and 82.2\% F1 on Prost. These results highlight the effectiveness of both the Bi-SSM design and the unified hypergraph structure in capturing complex spatial-contextual relationships in WSIs.
% The ablation experiment on the HGMamba block evaluates the impact of various configurations on the F1 score using the Prost dataset. As shown in Fig.~\ref{ablation_experiments}(a), Bi-SSM outperforms the SSM baseline with an F1 score of 69.3 (vs. 65.3 for SSM). Combining GCN with SSM-based methods improves performance, with GCN + SSM achieving 75.3 and GCN + Bi-SSM reaching 78.4. HGMamba surpasses all configurations, achieving an F1 score of 82.4, benefiting from better graph construction and diverse scanning techniques. These results highlight HGMamba’s superior architectural and graph processing advancements.
% Bi-SSM outperforms SSM baseline, as evidenced by its higher F1 score of 69.3 compared to 65.3 for SSM. Combining GCN with SSM-based methods results in better performance than using them individually, with GCN + SSM achieving 75.3 and GCN + Bi-SSM reaching 78.4. However, HGMamba outperforms all configurations with an F1 score of 82.4, benefiting from improved graph construction strategies and a more diverse graph scanning scheme. These results emphasize that HGMamba leverages both better architectural design and enhanced graph processing techniques for superior performance.

\subsubsection{Ablation on Tile Number.} 
As shown in Fig.~\ref{ablation_experiments}(a) and (b), the experiments show that the computational complexity of Graph Transformer increases quadratically with the number of tiles (N), as demonstrated in both the FLOPs and GPU memory. This quadratic growth becomes prohibitive in Whole Slide Image (WSI) analysis, where tile counts often reach tens of thousands, making such models impractical for real-world deployment. 
In contrast, WSI-HGMamba exhibits a linear relationship between computational cost and the number of tiles, offering a clear advantage in pathology MIL tasks.

\subsubsection{Ablation on Hypergraph Construction.} 
% As shown in Fig.~\ref{ablation_experiments}(d), the ablation in hypergraph construction highlights the impact of different adjacency schemes on the AUC performance in four datasets. The rule-based adjacency method focuses on the pathological local context. However, the similarity-based adjacency approach enables the incorporation of long-range dependencies. The best results are obtained by combining both methods, showcasing the complementary strengths in capturing both local and global relationships.
As shown in Fig.~\ref{ablation_experiments}(d), the ablation study on hypergraph construction evaluates the impact of different adjacency schemes on AUC performance across four datasets. The rule-based adjacency captures the pathological local context by connecting spatially adjacent tiles, which helps preserve fine-grained tissue structure. In contrast, the similarity-based adjacency captures long-range semantic relationships by linking tiles with similar visual features regardless of spatial proximity. While each strategy individually improves performance, the best results are achieved by combining both, demonstrating their complementary strengths in modeling both local and global interactions. 
This unified design leads to more informative hyperedges, ultimately enhancing the discriminative power of the constructed hypergraph.

\subsubsection{Ablation on Hypergraph Scanning.} 
% Fig.~\ref{ablation_experiments}(e) shows the ablation study comparing hypergraph scanning methods. Hypergraph Depth-First Search (H-DFS) captures the structural connectivity of the hypergraph, making it well-suited for datasets where capturing hierarchical relationships is crucial. On the other hand, Hypergraph Acyclic Random Walk (H-ARW) allows for stochastic exploration of arbitrary regions within the hypergraph, enhancing the ability to discover complex relationships. The combination of both methods delivers the best AUC scores across datasets, demonstrating the benefits of integrating structural and exploratory scanning strategies. Notably, all methods show a significant improvement over the random node scanning method.
Fig.~\ref{ablation_experiments}(e) presents the ablation study comparing different hypergraph scanning strategies. Hypergraph Depth-First Search (H-DFS) effectively captures the underlying structural connectivity of the hypergraph, making it particularly suitable for datasets where preserving hierarchical or sequential tissue relationships is important. In contrast, Hypergraph Acyclic Random Walk (H-ARW) facilitates stochastic traversal across diverse regions of the hypergraph, enabling the discovery of non-local and complex dependencies. The combination of H-DFS and H-ARW achieves the highest AUC scores across all datasets, highlighting the advantage of jointly modeling structural integrity and exploratory diversity. Notably, all proposed methods outperform the random node scanning baseline by a significant margin, confirming the importance of principled scanning strategies in transforming relational features into sequences for effective modeling.

\subsubsection{Ablation on HGMamba Layer.} 
% As shown in Fig.~\ref{ablation_experiments}(f), the ablation study on HGMamba layers compares the GCN and HGMamba models in varying numbers of layers on the Prost dataset. The GCN model experiences diminishing returns as the number of layers increases, with the best performance observed at 2 layers. This decline is attributed to the over-smoothing~\cite{chen2020measuring,keriven2022not} issue, where deeper layers lead to less informative node representations. In contrast, the HGMamba model, which incorporates the SSM structure, mitigates over-smoothing and maintains steady improvement in F1 score across all layers, which demonstrates HGMamba’s robustness in handling deeper layers, consistently outperforming GCN.
As shown in Fig.~\ref{ablation_experiments}(f), the ablation study on HGMamba layers evaluates the performance of GCN and HGMamba models with varying layer depths on the Prost dataset. The GCN model exhibits diminishing returns as the number of layers increases, peaking at 2 layers. Beyond this point, performance degrades due to the well-known over-smoothing effect~\cite{chen2020measuring,keriven2022not}, where deeper GCN layers cause node features to become indistinguishable and less informative. In contrast, the HGMamba model, enhanced with a State Space Model (SSM) structure, effectively alleviates over-smoothing by introducing sequential dynamics during feature propagation. As a result, HGMamba maintains consistent performance gains with increased depth, achieving higher F1 scores across all settings, highlighting its robustness and scalability, making it better suited for deeper architectures in complex WSI analysis tasks.

\subsubsection{Ablation on Top-K Neighbors.}
As illustrated in Fig.~\ref{ablation_experiments}(c), we conduct an ablation study to investigate the impact of varying the Top-K neighbors in the hypergraph construction for WSI-HGMamba on the Prost dataset. The F1 score improves as $k$ increases from 1 to 3, reaching a peak at $k=3$ (82.2\%). However, further increasing $k$ leads to a gradual decline in performance. This phenomenon can be attributed to the inclusion of less relevant or noisy neighbors when $k$ is large, which may introduce redundant or irrelevant information into the hypergraph, thereby hampering model discrimination. The results demonstrate that a moderate Top-K selection (\textit{e.g.}, $k=3$) provides the best balance between local connectivity and noise, yielding optimal performance.

\section{Conclusion}
% The WSI-HGMamba framework introduces a novel approach to large-scale WSI analysis, combining the strengths of HGNNs and Mamba. By integrating relational modeling with computational efficiency, our method significantly reduces the computational complexity while achieving high performance. With up to a $7\times$ reduction in FLOPs compared to Transformer-based models, WSI-HGMamba offers a scalable solution to tackle the challenges of analyzing high-resolution tissue slides, positioning it as a promising backbone for the next generation of pathology AI diagnosis. Future work will focus on extending WSI-HGMamba to the cellular graph for pathological micro-environment modeling.
We present WSI-HGMamba, a novel framework that unifies the high-order relational modeling capacity of hypergraph neural networks with the efficient long-range dependency modeling of Mamba-based state space models for large-scale WSI analysis. WSI-HGMamba effectively captures both spatial and contextual relationships across ultra-high-resolution WSIs while maintaining linear computational complexity. Compared to Transformer and Graph Transformer counterparts, our approach achieves up to a $7\times$ reduction in FLOPs, significantly improving scalability without sacrificing accuracy. These results position WSI-HGMamba as a strong candidate for the next generation of efficient and expressive pathology AI models. In future work, we aim to extend this framework toward fine-grained cellular graphs and microenvironment-level representations, enabling a deeper understanding of tissue heterogeneity and pathological context.

\bibliography{aaai2026}


\end{document}